\documentclass{article}

\PassOptionsToPackage{numbers,compress}{natbib}

\usepackage[preprint,eandd]{neurips_2026}

\usepackage[utf8]{inputenc}
\usepackage[T1]{fontenc}
\usepackage{hyperref}
\usepackage{url}
\usepackage{booktabs}
\usepackage{amsfonts}
\usepackage{amsmath}
\usepackage{nicefrac}
\usepackage{microtype}
\usepackage{xcolor}
\usepackage{graphicx}
\usepackage{multirow}
\usepackage{makecell}
\usepackage{tikz}

\graphicspath{{figures/}}

\usepackage{enumitem}
\usepackage{tikz}
\usetikzlibrary{arrows.meta, positioning}
\usepackage{amsmath}
\usepackage{booktabs}
\usepackage{multirow}
\usepackage{makecell}
\usepackage{amssymb}

\usepackage{xcolor}
\usepackage{colortbl}
\usepackage{pifont}
\usepackage{tikz}

\usepackage[most]{tcolorbox}
\definecolor{findingbg}{RGB}{240,252,253}
\definecolor{findingborder}{RGB}{74,150,158}

\newtcolorbox{findingbox}{
    colback=findingbg,
    colframe=findingborder,
    boxrule=0.75pt,
    arc=3pt,
    left=6pt,
    right=6pt,
    top=4pt,
    bottom=4pt,
    before skip=6pt,
    after skip=8pt,
}

\newcommand{\gsw}{\textsc{3D-GSW}}

\title{Characterizing Detectability in 3DGS Poisoning: \\A Stage-wise Benchmark}

\author{%
  \textbf{Quoc-Anh Bui-Huynh}$^{1,2,3}$ \quad
  \textbf{Thanh Duc Ngo}$^{2,3}$ \quad
  \textbf{Xue Geng}$^{4}$ \quad
  \textbf{Kaixin Xu}$^{4}$ \\
  \textbf{Wang Zhe}$^{4}$ \quad
  \textbf{Xulei Yang}$^{4}$ \quad
  \textbf{Ngai-Man Cheung}$^{1}$\thanks{Corresponding author.} \\[0.5em]
  $^{1}$Temasek Laboratories, Singapore University of Technology and Design \\
  $^{2}$Vietnam National University, Ho Chi Minh City \quad
  $^{3}$University of Information Technology, VNU-HCM \\
  $^{4}$Agency for Science, Technology, and Research (A*STAR) \\[0.3em]
  \texttt{\{huynh\_bui, ngaiman\_cheung\}@sutd.edu.sg} \quad
  \texttt{thanhnd@uit.edu.vn} \\
  \texttt{\{geng\_xue, xu\_kaixin, zhe\_wang, yang\_xulei\}@a-star.edu.sg}
}

\begin{document}

\maketitle

\begin{abstract}



3D Gaussian Splatting (3DGS) has rapidly emerged as a leading representation for real-time novel view synthesis, but recent work has shown that it is vulnerable to diverse poisoning attacks, including illusory object injection, computation cost amplification, and post hoc model watermarking. Despite this expanding threat surface, existing studies primarily focus on attack success, while defense and detection remain underexplored. From a detection perspective, a key challenge and opportunity arise from the multi-stage nature of the 3DGS reconstruction pipeline, which produces heterogeneous intermediate representations. Importantly, forensic signals for detecting poisoning are inherently stage dependent: an attack introduced at one stage may produce signals that emerge only at later stages or become more detectable as the pipeline progresses. This motivates a {\em stage-wise view of detectability} that goes beyond single-stage evaluation.
{\bf In this work,}
we introduce \textbf{Poison-3DGS}, a benchmark designed for {\em stage-wise characterization of poisoning detection in 3DGS}. The benchmark exposes stage-specific artifacts, including multi-view images, geometry, training dynamics, and Gaussian parameters, and covers a diverse set of scenes and poisoning attacks. Using this benchmark, we conduct a systematic study of detectability across pipeline stages.
Our analysis reveals several key insights. First, detectability varies significantly across stages, and no single stage consistently dominates across attack types. Second, different attacks exhibit distinct stage-specific forensic signals, indicating that detection effectiveness 
depends critically on where signals are observed. 
Third, later stage signals such as training dynamics and Gaussian parameter statistics provide strong cues that are not observable at earlier stages, where the attack is introduced or remains less detectable.
Overall, our work provides a principled benchmark and the first systematic characterization of stage-dependent detectability in 3DGS, offering a foundation for future research on robust and reliable 3DGS systems.
{\bf We include the benchmark in the submission.}

\end{abstract}

%

\section{Introduction}
\label{sec:intro}

3D Gaussian Splatting (3DGS)~\cite{3dgs, chen2024survey} has recently emerged as a leading representation for real-time photorealistic scene reconstruction, with growing deployment in robotics, augmented and virtual reality, mapping, and content creation \cite{zhai2025splatloc, chen2024gaussianeditor, matsuki2024gaussian}. As 3DGS moves into production pipelines, the integrity of its training inputs and intermediate artifacts becomes a concrete security concern.

A growing number of poisoning attacks now target 3DGS under fundamentally different threat models. StealthAttack~\cite{stealthattack} composites an illusory object into a target view and injects Gaussian seed points into low-density voids of the Structure-from-Motion (SfM) point cloud \cite{COLMAP}. Poison-Splat~\cite{poisonsplat} adds bounded adversarial noise to training images and forces the optimizer to over-densify, inflating Gaussian count, GPU memory, and training time by a factor of two to seven. 
Meanwhile, 
3D-GSW~\cite{gsw3d} finetunes a trained model to embed a decodable covert watermark in Gaussian colour, opacity, rotation, and scale, while leaving positions and the training trajectory untouched. GuardSplat~\cite{guardsplat} similarly modifies a pre-trained model by embedding messages into spherical harmonic features while preserving geometry and rendering fidelity. Although originally proposed for watermarking or ownership protection, such post-training modifications alter the learned representation to encode hidden information. In this work, we adopt a broad definition of poisoning that includes any unauthorized intervention modifying the learned 3DGS representation.
These attacks therefore span data-level, training-level, and model-level interventions across the 3DGS pipeline, from input data to final model representation.

\textbf{Research Gap.} Existing 3DGS attack studies \cite{stealthattack, guardsplat, gsw3d, poisonsplat} focus on attacker-centric metrics, such as visual stealth, Gaussian inflation ratio, or watermark recovery accuracy, but lack a systematic, defender-centric evaluation of detectability. From a detection perspective, a key challenge and opportunity arise from the multi-stage nature of the 3DGS reconstruction pipeline. The process involves stage-specific artifacts, including multi-view images, geometry, training dynamics, and Gaussian parameters. As a result, forensic signals for detecting poisoning are inherently \emph{stage dependent}. Importantly, an attack introduced at one stage may produce an attack footprint whose corresponding forensic signal emerges only at later stages or becomes more detectable as the pipeline progresses. This multi-stage structure creates both challenges and opportunities for detection, as different stages expose distinct and complementary evidence, a property that is particularly pronounced in 3DGS compared to conventional data poisoning settings.

This motivates a deeper view of 3DGS poisoning detection: beyond asking whether detection is possible, it is important to study \emph{where} forensic signals arise and become detectable, and \emph{how} different stages contribute to detectability. While detection has been extensively studied in 2D image forensics~\cite{mvssnet, imlvit, sparsevit} and 3D anomaly detection~\cite{real3dad}, these settings typically assume a single representation or uniform access to data. In contrast, 3DGS is a multi-stage pipeline where detection performance depends on which stage is analyzed. Despite this, there is no unified framework or benchmark that enables systematic, stage wise characterization of forensic signals
or reveals detection effectiveness across different stages of the 3DGS pipeline.

\begin{figure*}[t]
    \centering
    \includegraphics[width=\textwidth]{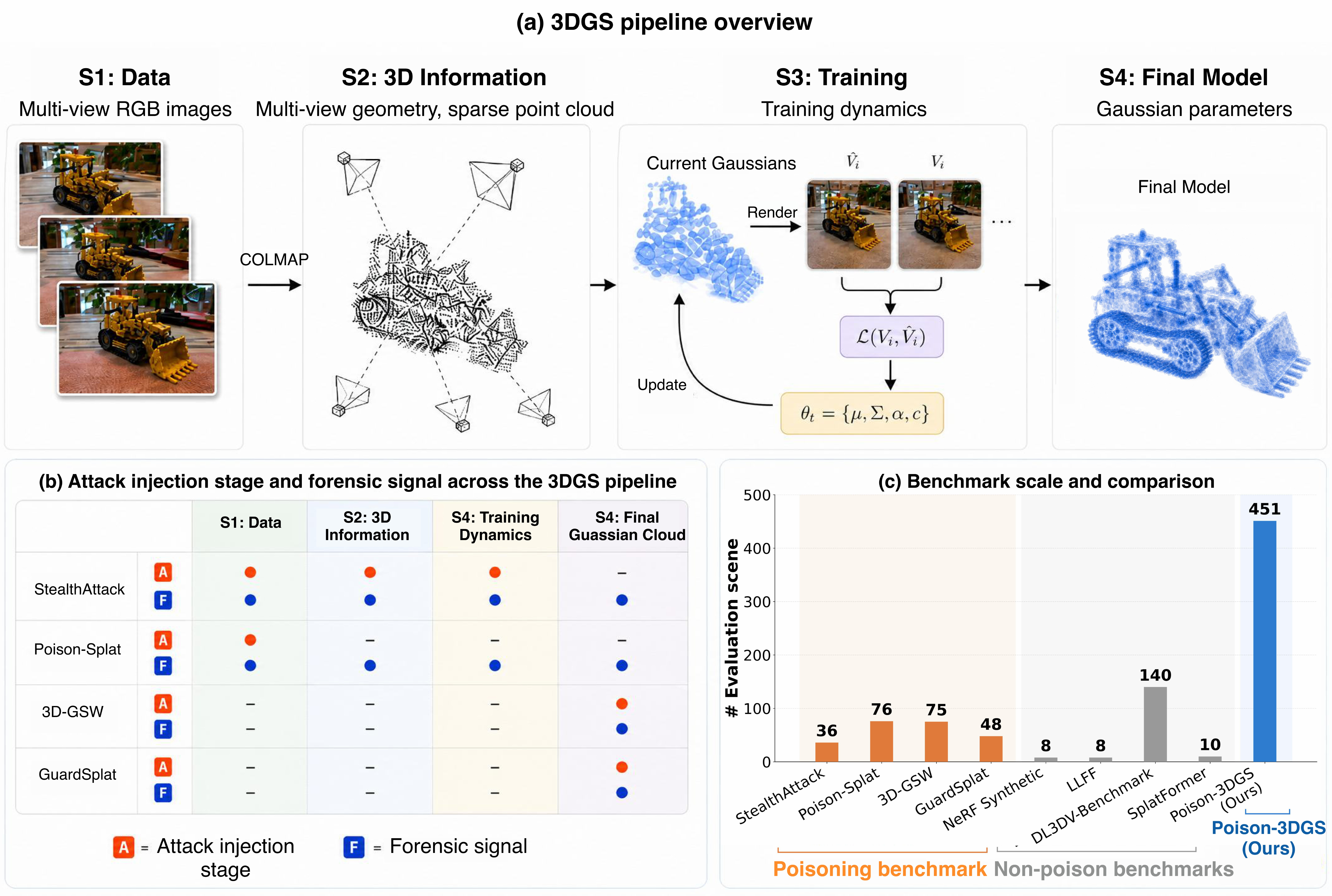}
\caption{{\bf Our stage-wise view of 3DGS poisoning detection.} 
(a) The multi-stage 3DGS reconstruction pipeline exposes {\em stage-specific artifacts}, including multi-view images, geometry, training dynamics, and Gaussian parameters. 
(b) {\em Attack injection stages} and their corresponding {\em forensic signals} across the pipeline. An attack introduced at one stage may produce its most detectable forensic signal at a different stage. 
(c) In addition to large scale and coverage, our benchmark supports  stage-wise characterization of detectability and 
analysis of distinct and complementary forensic signals at different stages.
The stage-wise formulation enables understanding of the contribution of individual sources of evidence.}
    \label{fig:pipeline}
\end{figure*}

{\bf In this paper}, we address this gap by introducing \textbf{Poison-3DGS}, a benchmark for studying 3DGS poisoning detection from a stage wise perspective. Building on the stage-dependent view illustrated in Fig.~\ref{fig:pipeline}, we move beyond treating detection as a single-stage problem and instead characterize how forensic signals arise and become informative across the 3DGS pipeline. To this end, Poison-3DGS is designed to expose stage-specific artifacts and enable systematic analysis of detectability across different stages. Using this benchmark, we conduct a comprehensive study across representative attacks and derive key insights into stage-dependent detection behavior. Our contributions are summarized as follows.


\begin{itemize}[nosep, leftmargin=3.em]

\item \textbf{Stage-wise characterization of forensic signals.}
We identify that in 3DGS poisoning, forensic signals are inherently stage-dependent due to the multi-stage reconstruction pipeline. An attack introduced at one stage may produce signals that emerge only at later stages or become more detectable as the pipeline progresses. This perspective reframes detection as understanding where informative signals arise across the pipeline.

\item \textbf{Benchmark grounded in stage-wise design.}
We construct \textbf{Poison-3DGS}, a benchmark that exposes stage-specific artifacts including multi-view images, geometry, training dynamics, and Gaussian parameters. This design enables controlled and systematic characterization of how different stages contribute to detectability, providing a principled framework beyond conventional single-stage evaluation.

\item \textbf{Empirical insights on stage-dependent detectability.}
Using Poison-3DGS, we show that detectability varies significantly across stages, and no single stage consistently dominates across attack types. In particular, later stage signals such as training dynamics and Gaussian parameter statistics provide strong cues that are not observable at earlier stages, where the attack is introduced or remains less detectable. This reveals new opportunities for detection.

\item \textbf{Stage-specific signatures of different attacks.}
We demonstrate that different attacks exhibit distinct stage-specific signatures. For example, StealthAttack is most detectable at data and training dynamic, Poison-Splat, 3D-GSW, and GuardSplat at the final model stage. These results highlight the importance of stage-aware analysis for understanding attack behavior.


\item \textbf{Stage-aware detector design and adaptation.}
We develop and adapt detectors to operate on stage-specific artifacts across the 3DGS pipeline, including multi-view images, geometry, training dynamics, and Gaussian parameters. Where suitable methods exist, we extend them to handle 3DGS-specific artifacts; where they do not, we design simple stage-specific detectors to capture relevant forensic signals. Our results show that off-the-shelf detectors transfer poorly across stages, while stage-aware designs that exploit stage-specific artifacts achieve more reliable detection.

\end{itemize}

\section{Related Work}
\label{sec:related}

\textbf{3DGS poisoning threats.}
Recent work shows that 3D Gaussian Splatting is vulnerable to attacks that directly affect users of
reconstructed scenes. StealthAttack~\cite{stealthattack} targets scene integrity by injecting
viewpoint-dependent illusory content through coordinated image and SfM manipulation, while
Poison-Splat~\cite{poisonsplat} targets availability by perturbing training images to trigger excessive
densification, increasing Gaussian count, memory use, and training cost. Although watermarking
methods such as \gsw~\cite{gsw3d} and GuardSplat~\cite{guardsplat} are designed for ownership
protection, they also demonstrate that a converged 3DGS model can be modified directly at the final
Gaussian representation to encode hidden information while preserving visual quality. GaussTrap~\cite{gausstrap} further shows that poisoned views can induce targeted scene
confusion. These studies establish a growing threat surface for 3DGS, but their evaluations remain
mostly attack-centric: each work studies its own scenes, artifacts, and success criteria. Poison-3DGS
instead studies these manipulations under a unified detector-centric protocol, with the goal of
characterizing where forensic signals become observable across the 3DGS pipeline.

\textbf{Defender-side evidence for 3DGS poisoning.}
Defender-side analysis of poisoning attacks is commonly framed as an evidence-discovery problem
~\cite{threatpoison,li2022backdoorlearningsurvey}: suspicious inputs may be detected through image
forensic cues~\cite{kadha2025unravelling}, poisoned samples may appear as representation outliers,
and abnormal training behavior may reveal attack-specific learning dynamics
~\cite{mvssnet,imlvit,Spectral,huang2025deep}. For 3DGS, this evidence-discovery problem is
naturally structured by the reconstruction pipeline, which exposes multiple artifact types, including
raw images, SfM geometry, optimization traces, and final Gaussian parameters. Recent 3DGS
defenses have begun to exploit parts of this pipeline, especially for computation-cost attacks:
RemedyGS detects and purifies poisoned input images, while Spectral Defense suppresses
attack-induced Gaussian growth through frequency-based filtering and regularization
~\cite{remedygs,spectraldefense}. These methods provide important mitigation mechanisms for
specific threat settings. Complementary to them, Poison-3DGS focuses on systematic evaluation; it provides a unified defender-centric protocol for characterizing where forensic signals
become observable across different attack families and stage-specific 3DGS artifacts.

\section{Preliminaries: 3D Gaussian Splatting}

3D Gaussian Splatting (3DGS) \cite{3dgs} represents a scene as a set of anisotropic Gaussian primitives
$\mathcal{G}=\{g_i\}_{i=1}^{N}$. Each Gaussian
$g_i=(\mu_i,\Sigma_i,\alpha_i,c_i)$ contains a 3D center $\mu_i$, covariance $\Sigma_i$,
opacity $\alpha_i$, and view-dependent color coefficients $c_i$. Given posed multi-view
images $\mathcal{D}=\{(V_k,\Pi_k)\}_{k=1}^{K}$, where $V_k$ is the ground-truth image and
$\Pi_k$ denotes the camera parameters, 3DGS renders a predicted view
$\hat V_k = \mathcal{R}(\mathcal{G};\Pi_k)$ through differentiable splatting and alpha
compositing.

Training optimizes the Gaussian parameters by minimizing a photometric reconstruction loss:
\begin{equation}
\min_{\mathcal{G}} \mathcal{L}_{\mathrm{photo}}(\mathcal{D})
=
\min_{\mathcal{G}}
\sum_{k=1}^{K}
\left[
(1-\lambda)\mathcal{L}_{1}(\hat V_k,V_k)
+
\lambda \mathcal{L}_{\mathrm{D\text{-}SSIM}}(\hat V_k,V_k)
\right],
\quad
\hat V_k=\mathcal{R}(\mathcal{G};\Pi_k),
\end{equation}
where $\lambda$ balances pixel-wise and structural reconstruction terms. During optimization,
3DGS also performs adaptive density control, which dynamically adds and removes Gaussian
primitives. Gaussians with large view-space positional gradients are densified through cloning or
splitting, while Gaussians with low opacity are pruned. This process allows 3DGS to allocate more
capacity to regions that remain difficult to reconstruct, but it also makes the number of Gaussians
and their parameter statistics part of the training dynamics.

\section{Poison-3DGS: Benchmarking Stage-wise Poisoning Detectability}
\label{sec:benchmark}

\subsection{A Defender-centric Benchmark Setup}



Existing 3DGS security works are primarily attack-centric: they evaluate whether an injected object appears, whether perturbations remain imperceptible, whether computation cost increases, or whether a watermark can be decoded. These metrics validate attack success, but they do not answer the defender-side question studied here: at which stage of the 3DGS pipeline can poisoning be detected? Prior poisoning and watermarking papers evaluate their own threat models in isolation, often with different scenes, artifacts, and success. Poison-3DGS instead turns heterogeneous 3DGS attacks into a unified stage-wise detection benchmark.

{\bf Stage-wise characterization of poisoning detection.}
Poison-3DGS studies poisoning detection under a unified detector-centric protocol over four
stage-specific artifact groups: \textbf{S1: Data}, raw multi-view RGB images; \textbf{S2: 3D
Information}, SfM cameras, tracks, and sparse points; \textbf{S3: Training Dynamics},
render--loss--update behavior including losses, gradients, densification, and Gaussian growth; and
\textbf{S4: Final Model}, trained Gaussian parameters and rendered outputs. Each detector operates
on the artifact group of a given stage and produces a scene-level anomaly score.

This stage-wise view is necessary because the \emph{attack injection stage} and the \emph{forensic signal}
are often decoupled. The attack injection stage is the artifact directly manipulated by the adversary, while
the forensic signal is the evidence that remains observable after the pipeline processes that
manipulation. For example, Poison-Splat perturbs images at \textbf{S1}, but its clearest signal appears
later through abnormal densification at \textbf{S3} and abnormal final Gaussian statistics at
\textbf{S4}. Conversely, post-training model-level attacks may leave no evidence in the input data,
SfM reconstruction, or training dynamics, but become observable only in the final Gaussian model.
Thus, Poison-3DGS evaluates not only whether a scene is poisoned, but also \emph{where} its forensic
signal becomes observable.

\noindent\textbf{Dataset.}
Poison-3DGS contains 37 clean scenes drawn from Free~\cite{f2nerf}, Mip-NeRF
360~\cite{mipnerf360}, and Tanks-and-Temples~\cite{tanksandtemples}. The scenes cover handheld
captures, indoor and outdoor environments, object-centric reconstructions, unbounded scenes, and
large architectural captures. For each scene, we provide raw images, SfM reconstruction, training
checkpoints, and a matched clean 3DGS reference model. Each poisoned variant is paired with its
clean counterpart, which makes scene-level binary detection well-defined and allows detectors to be
compared under the same scene distribution.

\begin{figure}
    \centering
    \includegraphics[width=1\linewidth]{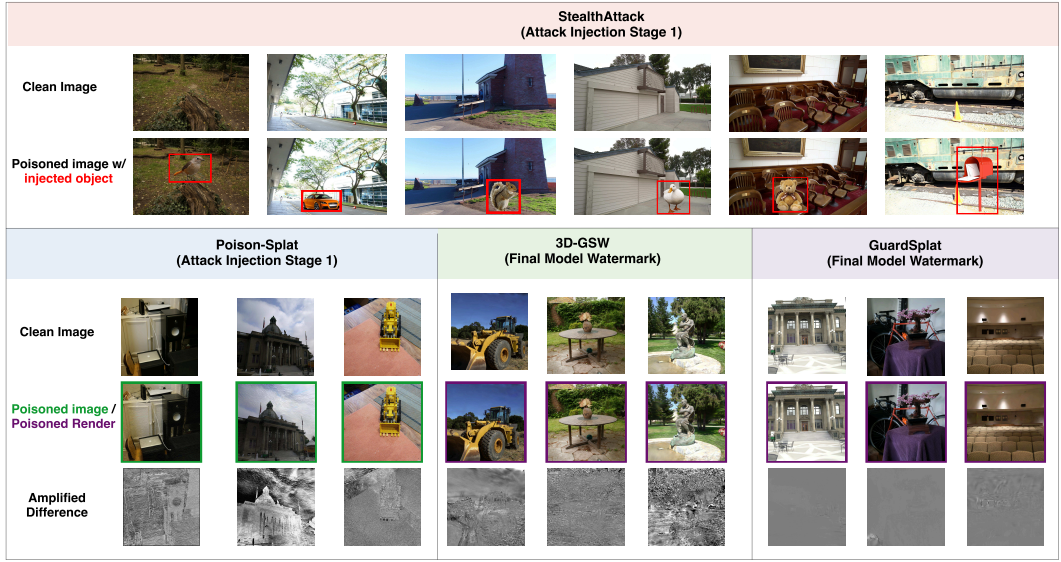}
    \caption{Qualitative examples from \textbf{Poison-3DGS}. StealthAttack shows \emph{image injection} by
    inserting an illusory object into a target training image, while Poison-Splat shows \emph{image
    perturbation} through subtle adversarial noise and its amplified residual. For \textbf{3D-GSW} and
    \textbf{GuardSplat}, we render the same sample view from the clean model and the watermarked
    model, and visualize amplified differences. The figure illustrates that different attack families in our
    benchmark leave distinct forensic signal at different stages of the 3DGS pipeline.}
    \label{fig:quanlitative}
\end{figure}

\noindent\textbf{Attack Space.}
Poison-3DGS, illustrated in Fig.~\ref{fig:quanlitative}, includes four attack families spanning
data-level, geometry-level, training-time, and final-model interventions. Rather than reproducing
each attack in a single isolated setting, we expand them into controlled variants that expose factors
relevant to detectability. For StealthAttack~\cite{stealthattack}, we scale the illusory-object setting
to 30 object identities and vary object size and target-view difficulty, yielding 120 variants. For
Poison-Splat~\cite{poisonsplat}, we vary the perturbation budget and poisoned-view ratio, including
practical partial-poisoning settings. For 3D-GSW~\cite{gsw3d} and GuardSplat~\cite{guardsplat}, we
vary message length and embedding strength while preserving their distinct final-model
interventions: 3D-GSW modifies broader Gaussian attributes, whereas GuardSplat concentrates its
changes in spherical-harmonic coefficients. Overall, Poison-3DGS contains 414 poisoned variants,
enabling controlled analysis of how attack surfaces and attack parameters shape stage-specific
forensic signals. See Appendix~\ref{app:additional_qualitative} for more examples.

\noindent\textbf{Metrics and Protocol.}
We formulate poisoning detection as binary scene-level detection, with clean scenes as negatives
and poisoned variants as positives. Each detector assigns a real-valued anomaly score to a scene at
a given stage, and scenes are ranked by this score for evaluation. Following score-based detector
evaluation protocols in OOD and anomaly detection \cite{hendrycks2016baseline, lee2018simple, yang2022openood}, we report AUROC to measure
overall separability between clean and poisoned scenes, and FPR@95TPR to measure the false-alarm
rate when detecting $95\%$ of poisoned variants. When a detector produces image-level, view-level,
point-level, checkpoint-level, or Gaussian-level scores, we aggregate them into a single scene-level
score using the detector-specific aggregation rule. Clean and poisoned variants of the same scene use
matched evaluation views, so detector scores and reconstruction metrics are computed on comparable
observations. For StealthAttack and Poison-Splat, we use a fixed 3DGS training schedule and record
per-iteration diagnostics; for 3D-GSW and GuardSplat, we provide the modified final Gaussian models,
embedding logs, and rendered outputs. We report both per-attack results and pooled results across
attack families.

\subsection{A Detector Suite for Stage-specific Forensic Signals}

To evaluate poisoning detectability across the 3DGS pipeline, we construct an unsupervised detector suite matched to the artifact group exposed at each stage. For fairness, all detectors use official pretrained weights or released off-the-shelf checkpoints, with no poisoned-sample fine-tuning, attack-specific adaptation, or access to attack labels. Each detector receives only its corresponding stage artifact and produces scene-level anomaly scores via stage-appropriate aggregation.

\noindent\textbf{S1: Data.}
At the data stage, detectors operate on the raw multi-view training images
$\mathcal{V}=\{V_k\}_{k=1}^{K}$. A data-stage attack modifies a subset of views $\mathcal{P}$ and
produces manipulated images $V'_k=\mathcal{A}(V_k)$ for $k\in\mathcal{P}$, such as the
target-view insertion in StealthAttack or the adversarial perturbations in Poison-Splat. We evaluate
MVSS-Net~\cite{mvssnet}, IML-ViT~\cite{ma2024imlvitbenchmarkingimagemanipulation}, and
SparseViT~\cite{sparsevit} by applying each detector $f_{\psi}$ independently
to every training view, yielding a view-level score $s_k=f_{\psi}(V_k)$, or
$s'_k=f_{\psi}(V'_k)$ for a manipulated view. Since poisoning can affect a small subset of views, we
aggregate view-level scores by max pooling, $S_{\mathrm{scene}}=\max_k s_k$. We also include
SoftPatch+~\cite{WANG2025111295}, which estimates patch-level outlier scores and calibrates them
using images from the same scene before producing a scene-level score. These baselines test whether
poisoning is already visible as image-level forensic signals before SfM reconstruction or 3DGS
optimization.

\noindent\textbf{S2: 3D Information.}
At the 3D information stage, we ask whether manipulated views become observable after SfM
reconstruction. Given the input views, COLMAP recovers camera intrinsics/extrinsics, feature tracks,
and a sparse point cloud $\mathcal{X}=\{x_j\}_{j=1}^{N}$ formed by triangulated 3D points. This
stage exposes two complementary forensic signals. First, manipulated views may violate multi-view
consistency: an inserted object may be supported only by the target view, while adversarial
perturbations may disturb feature matching, correspondence, or view-consistent geometry. We
therefore use MET3R~\cite{Asim_2025_CVPR}, MV-DUSt3R+~\cite{Tang_2025_CVPR},
NoPoSplat~\cite{ye2024poseproblemsurprisinglysimple}, and VGGT~\cite{Wang_2025_CVPR} as
multi-view consistency probes. Their reconstruction or consistency error is used as a geometric
anomaly score $s_{\mathrm{geo}}$ and aggregated into a scene-level score using the same max-pooling
strategy as S1. Second, injection-based attacks can leave sparse-geometry anomalies; for example,
StealthAttack may introduce artificial points along the object ray, creating abnormal local density
or geometry in $\mathcal{X}$. To capture this signal, we extract per-point features from coordinates,
RGB colors, and estimated normals using Sonata~\cite{Wu_2025_CVPR} or
Concerto~\cite{zhang2025concerto}, then fit an Isolation Forest \cite{liu2008isolation} within each scene to score local
geometric irregularities.

\noindent\textbf{S3: Training Dynamics.}
At the training stage, detectors operate on optimization diagnostics from saved 3DGS states
$\theta_t=\{g_i^t\}_{i=1}^{M_t}$, where the number of Gaussians $M_t$ changes over training due to densification and pruning. Unlike fixed-network gradient monitoring, this setting has a dynamic parameter set, no shared layer-wise embedding space across scenes, and view-dependent supervision: each training view affects only the Gaussians visible in its camera frustum. Nevertheless, poisoning can leave distinctive optimization fingerprints. For example, StealthAttack biases early optimization toward reconstructing the target-view illusion, while Poison-Splat is a DoS-style attack that induces excessive Gaussian growth and GPU memory cost. We therefore construct gradient-based baselines adapted to 3DGS dynamics. At each saved optimization state $\theta_t$, we summarize per-Gaussian update magnitudes, such as the positional-gradient norm $|\nabla_{x_i}\mathcal{L}(\theta_t)|_2$. We then aggregate these magnitudes across the visible Gaussians within each training view, yielding a view-indexed gradient matrix that captures how strongly each view drives the optimization. From this matrix, we derive scene-level anomaly scores using two unsupervised baselines: a GradNorm-style score over heavy-tailed gradient magnitudes~\cite{GradNorm}, and a Spectral-style score over the leading singular component of this matrix~\cite{Spectral}. These baselines test whether poisoning can be detected from optimization behavior before inspecting the final Gaussian representation.

\noindent\textbf{S4: Final Model.}
At the final-model stage, detectors operate on the optimized Gaussian representation
$\mathcal{G}^{\star}=\{g_i\}_{i=1}^{M}$, where each Gaussian stores geometry and appearance
attributes. This stage is security-critical because $\mathcal{G}^{\star}$ is the artifact ultimately
rendered, shared, or deployed, and it may carry forensic signals from input- or training-stage attacks
(e.g., StealthAttack's illusory object or Poison-Splat's over-densified Gaussian cloud) as well as
post-training modifications from 3D-GSW or GuardSplat. We use SceneSplat~\cite{Li_2025_ICCV},
SplatFormer~\cite{chen2024splatformer}, and Gaussian-MAE~\cite{ma2025large} as frozen Gaussian
encoders, providing complementary feature spaces through scene-level representation learning,
Gaussian refinement, and masked Gaussian-attribute reconstruction. For each scene, the final Gaussian cloud $\mathcal{G}^{\star}_i$ is independently encoded into a normalized scene embedding $\hat z_i$. We summarize the clean reference set by its normalized centroid $\hat\mu$, and score each scene by its deviation from this clean prototype using negative cosine similarity, $s_i=-\hat z_i^{\top}\hat\mu_{(i)}$, with a leave-one-out centroid when scoring a clean reference scene. This produces a scene-level anomaly score in the learned Gaussian feature space, where larger values indicate greater deviation from the clean final-model distribution.

\section{Analysis: Stage-specific Detectability and Detector Alignment}
\label{sec:analysis}

\begin{table*}[t]
\centering
\caption{
Stage-wise detector performance across poisoning attacks. Rows are grouped by the 3DGS pipeline
stage at which each detector operates. Entries report scene-level AUROC and FPR@95. Overall pools
the attacks evaluated at that stage. ``N/A'' indicates that the attack does not expose forensic signals at
that stage or that the required artifact is structurally unavailable. Bold denotes the best result
for each attack-specific column among applicable methods.
}
\label{tab:stagewise_detector_results}
\resizebox{\textwidth}{!}{
\begin{tabular}{l cc cc cc cc cc}
\toprule
\multirow{2}{*}{\textbf{Methods}} &
\multicolumn{2}{c}{\textbf{Overall}} &
\multicolumn{2}{c}{\textbf{StealthAttack}} &
\multicolumn{2}{c}{\textbf{Poison-Splat}} &
\multicolumn{2}{c}{\textbf{3D-GSW}} &
\multicolumn{2}{c}{\textbf{GuardSplat}} \\
\cmidrule(lr){2-3}
\cmidrule(lr){4-5}
\cmidrule(lr){6-7}
\cmidrule(lr){8-9}
\cmidrule(lr){10-11}
&
\textbf{AUROC}$\uparrow$ &
\textbf{FPR@95}$\downarrow$ &
\textbf{AUROC}$\uparrow$ &
\textbf{FPR@95}$\downarrow$ &
\textbf{AUROC}$\uparrow$ &
\textbf{FPR@95}$\downarrow$ &
\textbf{AUROC}$\uparrow$ &
\textbf{FPR@95}$\downarrow$ &
\textbf{AUROC}$\uparrow$ &
\textbf{FPR@95}$\downarrow$ \\
\midrule

\multicolumn{11}{l}{\textit{\textbf{S1: Data} \quad multi-view images}} \\
IML-ViT~(arXiv'24)~\cite{ma2024imlvitbenchmarkingimagemanipulation}
& 0.591 & 0.892
& 0.666 & 0.892
& 0.496 & 0.892
& N/A & N/A
& N/A & N/A \\
MVSS-Net~(TPAMI'22)~\cite{mvssnet}
& 0.414 & 0.973
& 0.511 & 0.919
& 0.290 & 1.000
& N/A & N/A
& N/A & N/A \\
SparseViT~(AAAI'25)~\cite{sparsevit}
& 0.550 & 0.946
& 0.636 & 0.892
& 0.440 & 1.000
& N/A & N/A
& N/A & N/A \\
SoftPatch+~(PR'25)~\cite{WANG2025111295}
& 0.617 & 0.973
& 0.768 & \textbf{0.622}
& 0.423 & 1.000
& N/A & N/A
& N/A & N/A \\

\midrule
\multicolumn{11}{l}{\textit{\textbf{S2: 3D Information} \quad multi-view geometry}} \\
MET3R~(CVPR'25)~\cite{Asim_2025_CVPR}
& 0.585 & 0.838
& 0.578 & 0.811
& 0.594 & 0.892
& N/A & N/A
& N/A & N/A \\
MV-DUSt3R+~(CVPR'25)~\cite{Tang_2025_CVPR}
& 0.525 & 0.973
& 0.481 & 0.973
& 0.581 & 0.973
& N/A & N/A
& N/A & N/A \\
NoPoSplat~(ICLR'25)~\cite{ye2024poseproblemsurprisinglysimple}
& 0.429 & 0.973
& 0.480 & 1.000
& 0.365 & 0.946
& N/A & N/A
& N/A & N/A \\
VGGT~(CVPR'25)~\cite{Wang_2025_CVPR}
& 0.487 & 1.000
& 0.520 & 0.973
& 0.445 & 1.000
& N/A & N/A
& N/A & N/A \\

\midrule
\multicolumn{11}{l}{\textit{\textbf{S2: 3D Information} \quad sparse point cloud}} \\
Sonata~(CVPR'25)~\cite{Wu_2025_CVPR}
& 0.766 & 0.892
& 0.766 & 0.892
& N/A$^\ddagger$ & N/A$^\ddagger$
& N/A & N/A
& N/A & N/A \\
Concerto~(NeurIPS'25)~\cite{zhang2025concerto}
& 0.725 & 0.946
& 0.725 & 0.946
& N/A$^\ddagger$ & N/A$^\ddagger$
& N/A & N/A
& N/A & N/A \\

\midrule
\multicolumn{11}{l}{\textit{\textbf{S3: Training} \quad training dynamics}} \\
GradNorm-style$^\dagger$~(NeurIPS'21)~\cite{GradNorm}
& 0.687 & 0.919
& \textbf{0.780} & 0.784
& 0.538 & 0.919
& N/A & N/A
& N/A & N/A \\
Spectral-style$^\dagger$~(NeurIPS'18)~\cite{Spectral}
& 0.582 & 0.919
& 0.572 & 0.919
& 0.598 & 0.919
& N/A & N/A
& N/A & N/A \\

\midrule
\multicolumn{11}{l}{\textit{\textbf{S4: Final Model} \quad Gaussian parameters}} \\
SceneSplat~(ICCV'25)~\cite{Li_2025_ICCV}
& 0.528 & 0.973
& 0.591 & 0.865
& 0.492 & 0.973
& 0.568 & 0.973
& 0.426 & 0.973 \\
SplatFormer~(ICLR'25)~\cite{chen2024splatformer}
& 0.619 & 0.946
& 0.681 & 0.946
& 0.594 & 0.784
& \textbf{0.689} & \textbf{0.919}
& 0.478 & 0.973 \\
Gaussian-MAE~(3DV'25)~\cite{ma2025large}
& 0.601 & 0.919
& 0.568 & 0.919
& \textbf{0.806} & \textbf{0.486}
& 0.570 & 0.919
& \textbf{0.488} & \textbf{0.973} \\

\bottomrule
\end{tabular}
}
\vspace{0.3em}

\begin{minipage}{\textwidth}
\footnotesize
$^\dagger$ These training-stage baselines are adapted to 3DGS training dynamics rather than used as direct classifier-level detectors. We report S3 results at the fixed checkpoint training iteration for all attacks and the overall split.
$^\ddagger$ Sparse point-cloud detectors are evaluated only when the poisoned artifact affects or exposes the SfM point cloud. For Poison-Splat, the original setup uses sparse point clouds generated from clean images, so this artifact is structurally unavailable for detecting the attack.
\end{minipage}
\end{table*}


\label{sec:analysis}

We analyze Poison-3DGS from a defender-centric perspective: where poisoning becomes detectable
in the 3DGS pipeline, and which detectors capture the corresponding forensic signals. Rather than
treating detection as a single decision over a completed scene, we study how forensic signals appear
across stage-specific artifacts, how they differ across attack families, and how detector performance
depends on the artifact being analyzed. Each detector follows our stage-specific access protocol:
it only uses artifacts available at its corresponding 3DGS stage and outputs a scene-level anomaly
score.
\subsection{Stage-specific Detectability Across Attacks}

\noindent\textbf{Where do forensic signals become most detectable?}
Table~\ref{tab:stagewise_detector_results} shows that detectability varies substantially across
stages, and no single stage consistently dominates across attack families. This supports a stage-wise
view of 3DGS poisoning detection: the most informative forensic signal depends on how each attack
interacts with the reconstruction pipeline.

StealthAttack leaves forensic signals across multiple stages. Its target-view insertion creates
localized image evidence at S1, where the best image-stage detector reaches 0.768 AUROC. The
injected geometry can also affect sparse 3D structure at S2. However, the strongest separation in our study appears at S3, where training
dynamics reach 0.780 AUROC. This is consistent with the attack mechanism: the attack-specific
noise scheduler perturbs clean-view supervision during training, making optimization less stable
across views. As a result, the training trajectory exhibits elevated per-view loss variability and
abnormal Gaussian update behavior.

\begin{figure}
    \centering
    \includegraphics[width=1\linewidth]{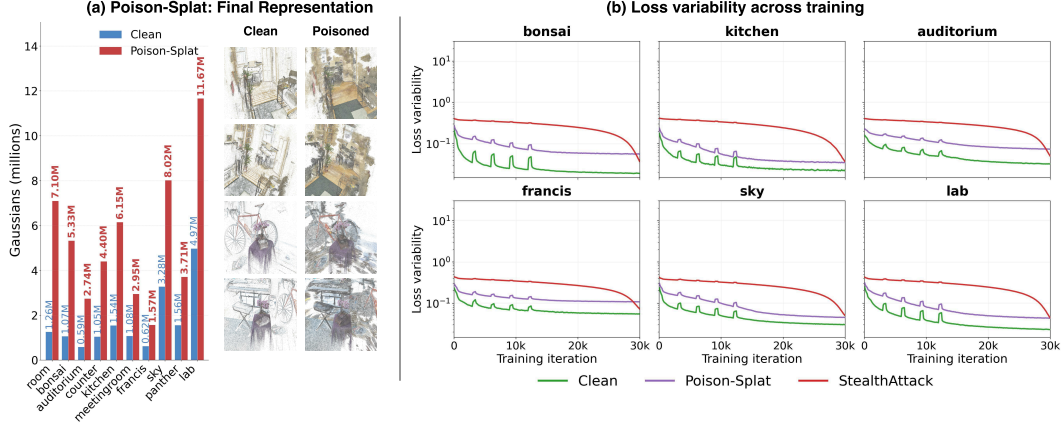}
    \caption{
    {\bf Stage-specific forensic signals for different attacks.}
    (a) Poison-Splat is most visible in the \emph{final Gaussian representation}: poisoned models contain
    more Gaussians and show denser Gaussian-center distributions than matched clean models, revealing
    \emph{over-densification} rather than render-quality differences.
    (b) StealthAttack is most visible in \emph{training dynamics}: its loss variability remains higher
    throughout optimization, indicating unstable view-wise supervision.
    These examples show that different attacks become most separable from different 3DGS artifacts.
    }
    \label{fig:detectability}
\end{figure}

Poison-Splat exhibits a different detectability profile. Although it is injected at S1 through bounded
image perturbations, these perturbations are weakly visible at the image stage. Their effect is amplified later through 3DGS optimization. By interfering with adaptive
density control, Poison-Splat induces excessive Gaussian growth, making the final Gaussian
representation the most informative stage for detection. At S4, Gaussian-MAE reaches 0.806 AUROC,
substantially higher than the image-stage result. As shown in Fig.~\ref{fig:detectability}, this appears
as increased Gaussian count and denser Gaussian-center distributions.

Post-training attacks form a third case. Since 3D-GSW and GuardSplat modify the trained Gaussian
representation after clean reconstruction, their forensic signals are only observable at S4 under our
stage-specific access protocol. The current S4 detectors provide uneven separation: 3D-GSW reaches
0.689 AUROC with SplatFormer, while GuardSplat remains difficult, with the best AUROC only
0.488. This suggests that final-model access is necessary for these attacks, but not sufficient unless
the detector is aligned with the specific Gaussian attributes being modified.

\begin{findingbox}
\textbf{\emph{Finding 1:}} Detectability varies significantly across stages, and no single stage
consistently dominates across attack families.
\end{findingbox}

\subsection{Forensic Signals Can Emerge After Attack Injection}

\noindent\textbf{Does the attack injection stage determine the detection stage?}
In a multi-stage 3DGS pipeline, the artifact directly modified by the adversary is not always the
artifact that provides the strongest forensic signal. Downstream processing can transform an early
manipulation into later-stage evidence through SfM reconstruction, optimization, or final Gaussian
parameterization. Therefore, evaluating only the attack injection stage may miss where poisoning
becomes most observable.

Poison-Splat and StealthAttack illustrate this decoupling. Poison-Splat is injected at S1 through
image perturbations, but its image-stage separability remains weak, with the best S1 AUROC reaching
only 0.496; its effect instead accumulates through optimization and becomes clearer as Gaussian
over-densification at S4. StealthAttack is introduced through target-view image and geometry
manipulation, making S1 and S2 informative, but its attack-specific training schedule further disrupts
view-wise optimization, making S3 the most separable stage in our detector suite. These cases show
that 3DGS poisoning should be analyzed as a pipeline process, not only at the point of attack injection.

\begin{findingbox}
\textbf{\emph{Finding 2:}} Attack injection and strongest detectability are not always aligned.
\end{findingbox}

\subsection{Detector Alignment with Stage-specific Forensic Signals}

\noindent\textbf{Which detectors capture the relevant forensic signals, and why?}
Table~\ref{tab:stagewise_detector_results} shows that existing detectors transfer unevenly across
stage-specific 3DGS artifacts. This unevenness reflects whether each detector is aligned with the
forensic signal exposed at its stage. For StealthAttack, SoftPatch+ is effective at S1 because the
attack edits only a few views and the inserted object is spatially localized; patch-level, scene-calibrated
scoring preserves this local evidence better than a global image score. At S3, the GradNorm-style
detector performs best because StealthAttack's noise scheduler perturbs clean-view supervision during
training, producing unstable view-wise optimization and abnormal per-Gaussian update magnitudes.

For Poison-Splat, Gaussian-MAE performs best among S4 detectors because its pretext task is closest
to the attack's final-model forensic signal. Poison-Splat mainly changes the distribution of Gaussian
primitives after abnormal densification, including count, density, scale, opacity, and geometry, rather
than semantic scene content. Gaussian-MAE learns from Gaussian splats through masked reconstruction
of primitive attributes, making it more sensitive to such local parameter changes. In contrast,
SceneSplat is designed for 3DGS scene understanding with vision-language pretraining, while
SplatFormer targets feed-forward refinement of Gaussian splats for robust novel-view rendering;
these objectives are less directly aligned with detecting abnormal primitive distributions.
Consistently, Table~\ref{tab:stagewise_detector_results} shows that Gaussian-MAE reaches 0.806
AUROC on Poison-Splat, compared with 0.492 for SceneSplat and 0.594 for SplatFormer.
SceneSplat’s focus on 3DGS scene understanding with vision-language pretraining and SplatFormer’s
focus on refinement for robust novel-view synthesis support this interpretation. 

The same alignment principle explains the different outcomes for 3D-GSW and GuardSplat. Both are
post-training attacks observable only at S4 under our protocol, but they modify different Gaussian
attributes. 3D-GSW applies a broader intervention over Gaussian geometry and appearance, producing
a more detectable final-representation shift, with SplatFormer reaching 0.689 AUROC. GuardSplat is
narrower: its signal is concentrated in spherical-harmonic appearance coefficients while preserving
geometry and rendering quality, so current S4 detectors remain weak, with the best AUROC only
0.488. Thus, final-model access is necessary but not sufficient; detectors must also match the specific
Gaussian attributes modified by the attack.

\begin{findingbox}
\textbf{\emph{Finding 3:}} Existing detectors transfer unevenly across stage-specific 3DGS artifacts.
\end{findingbox}

\section{Conclusion and Discussion}
\label{sec:conclusion}

This paper studies poisoning detection in 3D Gaussian Splatting from a stage-wise, defender-centric
perspective. While recent attacks manipulate input views, SfM geometry, training dynamics, and final
Gaussian parameters, existing evaluations mainly focus on attack success rather than where evidence
becomes detectable. We introduce \textbf{Poison-3DGS}, a benchmark for characterizing poisoning
detection across 3DGS pipeline stages, exposing artifacts from multi-view images, geometry, training
traces, and Gaussian parameters. Our study shows that detectability varies substantially across stages,
no single stage consistently dominates across attack types, and attack injection and strongest
detectability are often decoupled. We further find that detector effectiveness depends on alignment
with stage-specific forensic signals, suggesting that 3DGS security should be treated as a
pipeline-level problem rather than a single-artifact detection task. We hope Poison-3DGS provides a
foundation for future work on robust and reliable 3DGS systems.

\paragraph{Limitations and future directions.}
Poison-3DGS is built on the current public landscape of 3DGS attacks, using official or reproducible
implementations whenever available. As new attacks emerge, the benchmark can be extended while
preserving its core design: matched clean references, stage-wise artifacts, and scene-level detector
evaluation.

\begin{enumerate}[leftmargin=1.5em]
    \item \textbf{Extending the attack suite.}
    Future extensions can incorporate new poisoning, watermarking, or model-manipulation attacks by
    applying them to the same clean scene pool, recording their pipeline artifacts, and evaluating them
    under the same unified detector protocol. This would allow new threats to be compared by both
    attack success and where their forensic signals emerge across the 3DGS pipeline.

    \item \textbf{From detection to mitigation.}
    Poison-3DGS characterizes where forensic signals become observable, but does not propose a
    complete mitigation method. This motivates stage-aware defenses, such as filtering suspicious
    views, repairing SfM geometry, regularizing abnormal densification, or auditing final Gaussian
    attributes before deployment.
\end{enumerate}

\bibliographystyle{plainnat}
\bibliography{references}

@article{3dgs,
  title={3d gaussian splatting for real-time radiance field rendering.},
  author={Kerbl, Bernhard and Kopanas, Georgios and Leimk{\"u}hler, Thomas and Drettakis, George and others},
  journal={ACM Trans. Graph.},
  volume={42},
  number={4},
  pages={139--1},
  year={2023}
}

@inproceedings{stealthattack,
  title={StealthAttack: Robust 3D Gaussian Splatting Poisoning via Density-Guided Illusions},
  author={Ke, Bo-Hsu and Xie, You-Zhe and Liu, Yu-Lun and Chiu, Wei-Chen},
  booktitle={Proceedings of the IEEE/CVF International Conference on Computer Vision},
  pages={27400--27411},
  year={2025}
}

@article{poisonsplat,
  title={Poison-splat: Computation cost attack on 3d gaussian splatting},
  author={Lu, Jiahao and Zhang, Yifan and Shen, Qiuhong and Wang, Xinchao and Yan, Shuicheng},
  journal={arXiv preprint arXiv:2410.08190},
  year={2024}
}

@inproceedings{gsw3d,
  title={3d-gsw: 3d gaussian splatting for robust watermarking},
  author={Jang, Youngdong and Park, Hyunje and Yang, Feng and Ko, Heeju and Choo, Euijin and Kim, Sangpil},
  booktitle={Proceedings of the Computer Vision and Pattern Recognition Conference},
  pages={5938--5948},
  year={2025}
}

@inproceedings{mipnerf360,
  title={Mip-nerf 360: Unbounded anti-aliased neural radiance fields},
  author={Barron, Jonathan T and Mildenhall, Ben and Verbin, Dor and Srinivasan, Pratul P and Hedman, Peter},
  booktitle={Proceedings of the IEEE/CVF conference on computer vision and pattern recognition},
  pages={5470--5479},
  year={2022}
}

@article{tanksandtemples,
  title={Tanks and temples: Benchmarking large-scale scene reconstruction},
  author={Knapitsch, Arno and Park, Jaesik and Zhou, Qian-Yi and Koltun, Vladlen},
  journal={ACM Transactions on Graphics (ToG)},
  volume={36},
  number={4},
  pages={1--13},
  year={2017},
  publisher={ACM New York, NY, USA}
}

@inproceedings{f2nerf,
  title={F2-nerf: Fast neural radiance field training with free camera trajectories},
  author={Wang, Peng and Liu, Yuan and Chen, Zhaoxi and Liu, Lingjie and Liu, Ziwei and Komura, Taku and Theobalt, Christian and Wang, Wenping},
  booktitle={Proceedings of the IEEE/CVF Conference on Computer Vision and Pattern Recognition},
  pages={4150--4159},
  year={2023}
}

@article{mvssnet,
  title={Mvss-net: Multi-view multi-scale supervised networks for image manipulation detection},
  author={Dong, Chengbo and Chen, Xinru and Hu, Ruohan and Cao, Juan and Li, Xirong},
  journal={IEEE Transactions on Pattern Analysis and Machine Intelligence},
  volume={45},
  number={3},
  pages={3539--3553},
  year={2022},
  publisher={IEEE}
}

@inproceedings{imlvit,
  title={Multi-view feature extraction via tunable prompts is enough for image manipulation localization},
  author={Liu, Xuntao and Yang, Yuzhou and Wang, Haoyue and Ying, Qichao and Qian, Zhenxing and Zhang, Xinpeng and Li, Sheng},
  booktitle={Proceedings of the 32nd ACM International Conference on Multimedia},
  pages={9999--10007},
  year={2024}
}

@inproceedings{sparsevit,
  title={Can we get rid of handcrafted feature extractors? sparsevit: Nonsemantics-centered, parameter-efficient image manipulation localization through spare-coding transformer},
  author={Su, Lei and Ma, Xiaochen and Zhu, Xuekang and Niu, Chaoqun and Lei, Zeyu and Zhou, Ji-Zhe},
  booktitle={Proceedings of the AAAI conference on artificial intelligence},
  volume={39},
  number={7},
  pages={7024--7032},
  year={2025}
}

@inproceedings{real3dad,
  title={R3d-ad: Reconstruction via diffusion for 3d anomaly detection},
  author={Zhou, Zheyuan and Wang, Le and Fang, Naiyu and Wang, Zili and Qiu, Lemiao and Zhang, Shuyou},
  booktitle={European conference on computer vision},
  pages={91--107},
  year={2024},
  organization={Springer}
}

@inproceedings{COLMAP,
    author={Sch\"{o}nberger, Johannes Lutz and Frahm, Jan-Michael},
    title={Structure-from-Motion Revisited},
    booktitle={Conference on Computer Vision and Pattern Recognition (CVPR)},
    year={2016},
}

@misc{gausstrap,
      title={GaussTrap: Stealthy Poisoning Attacks on 3D Gaussian Splatting for Targeted Scene Confusion}, 
      author={Jiaxin Hong and Sixu Chen and Shuoyang Sun and Hongyao Yu and Hao Fang and Yuqi Tan and Bin Chen and Shuhan Qi and Jiawei Li},
      year={2025},
      eprint={2504.20829},
      archivePrefix={arXiv},
      primaryClass={cs.CV},
      url={https://arxiv.org/abs/2504.20829}, 
}

@misc{ma2024imlvitbenchmarkingimagemanipulation,
      title={IML-ViT: Benchmarking Image Manipulation Localization by Vision Transformer}, 
      author={Xiaochen Ma and Bo Du and Zhuohang Jiang and Xia Du and Ahmed Y. Al Hammadi and Jizhe Zhou},
      year={2024},
      eprint={2307.14863},
      archivePrefix={arXiv},
      primaryClass={cs.CV},
      url={https://arxiv.org/abs/2307.14863}, 
}

@article{WANG2025111295,
title = {SoftPatch+: Fully unsupervised anomaly classification and segmentation},
journal = {Pattern Recognition},
volume = {161},
pages = {111295},
year = {2025},
issn = {0031-3203},
doi = {https://doi.org/10.1016/j.patcog.2024.111295},
url = {https://www.sciencedirect.com/science/article/pii/S003132032401046X},
author = {Chengjie Wang and Xi Jiang and Bin-Bin Gao and Zhenye Gan and Yong Liu and Feng Zheng and Lizhuang Ma}
}

@InProceedings{Asim_2025_CVPR,
    author    = {Asim, Mohammad and Wewer, Christopher and Wimmer, Thomas and Schiele, Bernt and Lenssen, Jan Eric},
    title     = {MET3R: Measuring Multi-View Consistency in Generated Images},
    booktitle = {Proceedings of the IEEE/CVF Conference on Computer Vision and Pattern Recognition (CVPR)},
    month     = {June},
    year      = {2025},
    pages     = {6034-6044}
}

@InProceedings{Tang_2025_CVPR,
    author    = {Tang, Zhenggang and Fan, Yuchen and Wang, Dilin and Xu, Hongyu and Ranjan, Rakesh and Schwing, Alexander and Yan, Zhicheng},
    title     = {MV-DUSt3R+: Single-Stage Scene Reconstruction from Sparse Views In 2 Seconds},
    booktitle = {Proceedings of the IEEE/CVF Conference on Computer Vision and Pattern Recognition (CVPR)},
    month     = {June},
    year      = {2025},
    pages     = {5283-5293}
}

@misc{ye2024poseproblemsurprisinglysimple,
      title={No Pose, No Problem: Surprisingly Simple 3D Gaussian Splats from Sparse Unposed Images}, 
      author={Botao Ye and Sifei Liu and Haofei Xu and Xueting Li and Marc Pollefeys and Ming-Hsuan Yang and Songyou Peng},
      year={2024},
      eprint={2410.24207},
      archivePrefix={arXiv},
      primaryClass={cs.CV},
      url={https://arxiv.org/abs/2410.24207}, 
}

@InProceedings{Wang_2025_CVPR,
    author    = {Wang, Jianyuan and Chen, Minghao and Karaev, Nikita and Vedaldi, Andrea and Rupprecht, Christian and Novotny, David},
    title     = {VGGT: Visual Geometry Grounded Transformer},
    booktitle = {Proceedings of the IEEE/CVF Conference on Computer Vision and Pattern Recognition (CVPR)},
    month     = {June},
    year      = {2025},
    pages     = {5294-5306}
}

@InProceedings{Wu_2025_CVPR,
    author    = {Wu, Xiaoyang and DeTone, Daniel and Frost, Duncan and Shen, Tianwei and Xie, Chris and Yang, Nan and Engel, Jakob and Newcombe, Richard and Zhao, Hengshuang and Straub, Julian},
    title     = {Sonata: Self-Supervised Learning of Reliable Point Representations},
    booktitle = {Proceedings of the IEEE/CVF Conference on Computer Vision and Pattern Recognition (CVPR)},
    month     = {June},
    year      = {2025},
    pages     = {22193-22204}
}

@inproceedings{zhang2025concerto,
  title={Concerto: Joint 2D-3D Self-Supervised Learning Emerges Spatial Representations},
  author={Zhang, Yujia and Wu, Xiaoyang and Lao, Yixing and Wang, Chengyao and Tian, Zhuotao and Wang, Naiyan and Zhao, Hengshuang},
  booktitle={NeurIPS},
  year={2025}
}

@inproceedings{GradNorm,
author = {Huang, Rui and Geng, Andrew and Li, Yixuan},
title = {On the importance of gradients for detecting distributional shifts in the wild},
year = {2021},
isbn = {9781713845393},
publisher = {Curran Associates Inc.},
address = {Red Hook, NY, USA},
booktitle = {Proceedings of the 35th International Conference on Neural Information Processing Systems},
articleno = {52},
numpages = {13},
series = {NIPS '21}
}

@InProceedings{Li_2025_ICCV,
    author    = {Li, Yue and Ma, Qi and Yang, Runyi and Li, Huapeng and Ma, Mengjiao and Ren, Bin and Popovic, Nikola and Sebe, Nicu and Konukoglu, Ender and Gevers, Theo and Van Gool, Luc and Oswald, Martin R. and Paudel, Danda Pani},
    title     = {SceneSplat: Gaussian Splatting-based Scene Understanding with Vision-Language Pretraining},
    booktitle = {Proceedings of the IEEE/CVF International Conference on Computer Vision (ICCV)},
    month     = {October},
    year      = {2025},
    pages     = {4961-4972}
}

@misc{chen2024splatformer,
    title = {SplatFormer: Point Transformer for Robust 3D Gaussian Splatting},
    author = {Chen, Yutong and Mihajlovic, Marko and Chen, Xiyi and Wang, Yiming and Prokudin, Sergey and Tang, Siyu},
    booktitle = {International Conference on Learning Representations (ICLR)},
    year = {2025}
}

@inproceedings{ma2025large,
    title={A Large-Scale Dataset of Gaussian Splats and Their Self-Supervised Pretraining},
    author={Ma, Qi and Li, Yue and Ren, Bin and Sebe, Nicu and Konukoglu, Ender and Gevers, Theo and Van Gool, Luc and Paudel, Danda Pani},
    booktitle={2025 International Conference on 3D Vision (3DV)},
    pages={145--155},
    year={2025},
    organization={IEEE}
  }

@inproceedings{guardsplat,
  title={GuardSplat: efficient and robust watermarking for 3d gaussian splatting},
  author={Chen, Zixuan and Wang, Guangcong and Zhu, Jiahao and Lai, Jianhuang and Xie, Xiaohua},
  booktitle={Proceedings of the Computer Vision and Pattern Recognition Conference},
  pages={16325--16335},
  year={2025}
}

@article{Spectral,
  title={Spectral signatures in backdoor attacks},
  author={Tran, Brandon and Li, Jerry and Madry, Aleksander},
  journal={Advances in neural information processing systems},
  volume={31},
  year={2018}
}

@article{remedygs,
  title={RemedyGS: Defend 3D Gaussian Splatting against Computation Cost Attacks},
  author={Li, Yanping and Liu, Zhening and Li, Zijian and Lin, Zehong and Zhang, Jun},
  journal={arXiv preprint arXiv:2511.22147},
  year={2025}
}

@article{spectraldefense,
  title={Spectral Defense Against Resource-Targeting Attack in 3D Gaussian Splatting},
  author={Chen, Yang and Yu, Yi and He, Jiaming and Duan, Yueqi and Zhu, Zheng and Tan, Yap-Peng},
  journal={arXiv preprint arXiv:2603.12796},
  year={2026}
}

@article{chen2024survey,
  title={A survey on 3d gaussian splatting},
  author={Chen, Guikun and Wang, Wenguan},
  journal={ACM Computing Surveys},
  year={2024},
  publisher={ACM New York, NY}
}

@article{zhai2025splatloc,
  title={Splatloc: 3d gaussian splatting-based visual localization for augmented reality},
  author={Zhai, Hongjia and Zhang, Xiyu and Zhao, Boming and Li, Hai and He, Yijia and Cui, Zhaopeng and Bao, Hujun and Zhang, Guofeng},
  journal={IEEE Transactions on Visualization and Computer Graphics},
  year={2025},
  publisher={IEEE}
}

@inproceedings{chen2024gaussianeditor,
  title={Gaussianeditor: Swift and controllable 3d editing with gaussian splatting},
  author={Chen, Yiwen and Chen, Zilong and Zhang, Chi and Wang, Feng and Yang, Xiaofeng and Wang, Yikai and Cai, Zhongang and Yang, Lei and Liu, Huaping and Lin, Guosheng},
  booktitle={Proceedings of the IEEE/CVF conference on computer vision and pattern recognition},
  pages={21476--21485},
  year={2024}
}

@inproceedings{matsuki2024gaussian,
  title={Gaussian splatting slam},
  author={Matsuki, Hidenobu and Murai, Riku and Kelly, Paul HJ and Davison, Andrew J},
  booktitle={Proceedings of the IEEE/CVF conference on computer vision and pattern recognition},
  pages={18039--18048},
  year={2024}
}

@article{kadha2025unravelling,
  title={Unravelling digital forgeries: A systematic survey on image manipulation detection and localization},
  author={Kadha, VijayaKumar and Bakshi, Sambit and Das, Santos Kumar},
  journal={ACM Computing Surveys},
  volume={57},
  number={12},
  pages={1--36},
  year={2025},
  publisher={ACM New York, NY}
}

@article{huang2025deep,
  title={Deep learning advancements in anomaly detection: A comprehensive survey},
  author={Huang, Haoqi and Wang, Ping and Pei, Jianhua and Wang, Jiacheng and Alexanian, Shahen and Niyato, Dusit},
  journal={IEEE Internet of Things Journal},
  year={2025},
  publisher={IEEE}
}

@article{threatpoison,
author = {Wang, Zhibo and Ma, Jingjing and Wang, Xue and Hu, Jiahui and Qin, Zhan and Ren, Kui},
title = {Threats to Training: A Survey of Poisoning Attacks and Defenses on Machine Learning Systems},
year = {2022},
issue_date = {July 2023},
publisher = {Association for Computing Machinery},
address = {New York, NY, USA},
volume = {55},
number = {7},
issn = {0360-0300},
url = {https://doi.org/10.1145/3538707},
doi = {10.1145/3538707},
journal = {ACM Comput. Surv.},
month = dec,
articleno = {134},
numpages = {36}
}

@misc{li2022backdoorlearningsurvey,
      title={Backdoor Learning: A Survey}, 
      author={Yiming Li and Yong Jiang and Zhifeng Li and Shu-Tao Xia},
      year={2022},
      eprint={2007.08745},
      archivePrefix={arXiv},
      primaryClass={cs.CR},
      url={https://arxiv.org/abs/2007.08745}, 
}

@inproceedings{liu2008isolation,
  title={Isolation forest},
  author={Liu, Fei Tony and Ting, Kai Ming and Zhou, Zhi-Hua},
  booktitle={2008 eighth ieee international conference on data mining},
  pages={413--422},
  year={2008},
  organization={IEEE}
}

@article{hendrycks2016baseline,
  title={A baseline for detecting misclassified and out-of-distribution examples in neural networks},
  author={Hendrycks, Dan and Gimpel, Kevin},
  journal={arXiv preprint arXiv:1610.02136},
  year={2016}
}

@article{lee2018simple,
  title={A simple unified framework for detecting out-of-distribution samples and adversarial attacks},
  author={Lee, Kimin and Lee, Kibok and Lee, Honglak and Shin, Jinwoo},
  journal={Advances in neural information processing systems},
  volume={31},
  year={2018}
}

@article{yang2022openood,
  title={Openood: Benchmarking generalized out-of-distribution detection},
  author={Yang, Jingkang and Wang, Pengyun and Zou, Dejian and Zhou, Zitang and Ding, Kunyuan and Peng, Wenxuan and Wang, Haoqi and Chen, Guangyao and Li, Bo and Sun, Yiyou and others},
  journal={Advances in Neural Information Processing Systems},
  volume={35},
  pages={32598--32611},
  year={2022}
}

\end{document}